\documentclass[10pt,twocolumn,letterpaper]{article}

\usepackage{iccv}
\usepackage{times}
\usepackage{epsfig}
\usepackage{graphicx}
\usepackage{amsmath}
\usepackage{amssymb}
\usepackage{dsfont}
\usepackage{siunitx}
\usepackage{multirow}
\usepackage{multicol}
\usepackage{subcaption}
\usepackage{enumitem} 
\usepackage{listings}

\usepackage{xcolor}

\definecolor{codegreen}{rgb}{0,0.6,0}
\definecolor{codegray}{rgb}{0.501961,0.501961,0.501961}
\definecolor{codepurple}{rgb}{0.58,0,0.82}
\definecolor{codeblack}{rgb}{0,0,0}

\lstdefinestyle{mystyle}{
  backgroundcolor=\color{white},   
  commentstyle=\color{codegray},
  keywordstyle=\ttfamily\scriptsize\color{codeblack},
  numberstyle=\tiny\color{codegray},
  stringstyle=\color{codegreen},
  basicstyle=\ttfamily\scriptsize\color{codeblack},
  breakatwhitespace=false,         
  breaklines=true,                 
  captionpos=b,                    
  keepspaces=false,                 
  showspaces=false,                
  showstringspaces=false,
  showtabs=false,                  
  tabsize=2
}

\usepackage{algorithm}
\usepackage{array}

\makeatletter
\newcommand{\thickhline}{%
    \noalign {\ifnum 0=`}\fi \hrule height 1pt
    \futurelet \reserved@a \@xhline
}
\makeatother

\newcommand{\gray}[1]{\textcolor{gray}{#1}}
\newcommand{\algref}[1]{Algorithm~\ref{#1}}
\newcommand{\eqnref}[1]{Eqn.(\ref{#1})}
\newcommand{\figref}[1]{Fig.~\ref{#1}}
\newcommand{\secref}[1]{Section~\ref{#1}}
\newcommand{\tabref}[1]{Table~\ref{#1}}

\usepackage{amssymb}
\usepackage{pifont}
\newcommand{\algcomment}[1]{%
    \vspace{-\baselineskip}%
    \noindent%
    {\footnotesize #1\par}%
    \vspace{\baselineskip}%
}

\usepackage[pagebackref=true,breaklinks=true,letterpaper=true,colorlinks,bookmarks=false]{hyperref}

\iccvfinalcopy 

\def\iccvPaperID{****} 

\begin{document}

\title{UniMoCo: Unsupervised, Semi-Supervised and Full-Supervised \\ Visual Representation Learning}

\author{Zhigang Dai\textsuperscript{1*}, Bolun Cai\textsuperscript{2$\dagger$}, Yugeng Lin\textsuperscript{2}, Junying Chen\textsuperscript{1} \\ \textsuperscript{1}South China University of Technology   \textsuperscript{2}Tencent Wechat AI\\
{\tt\small zhigangdai@hotmail.com, \{arlencai,lincolnlin\}@tencent.com, jychense@scut.edu.cn}}

\maketitle
{\let\thefootnote\relax\footnote{{\textsuperscript{*}This work is done when Zhigang Dai was an intern at 
Tencent Wechat AI.}}}
{\let\thefootnote\relax\footnote{{\textsuperscript{$\dagger$}Corresponding author.}}}

\begin{abstract}
    
Momentum Contrast (MoCo) achieves great success for unsupervised visual representation. However, there are a lot of supervised and semi-supervised datasets, which are already labeled. To fully utilize the label annotations, we propose Unified Momentum Contrast (UniMoCo), which extends MoCo to support arbitrary ratios of labeled data and unlabeled data training. Compared with MoCo, UniMoCo has two modifications as follows: (1) Different from a single positive pair in MoCo, we maintain multiple positive pairs on-the-fly by comparing the query label to a label queue. (2) We propose a Unified Contrastive(UniCon) loss to support an arbitrary number of positives and negatives in a unified pair-wise optimization perspective. Our UniCon is more reasonable and powerful than the supervised contrastive loss in theory and practice. In our experiments, we pre-train multiple UniMoCo models with different ratios of ImageNet labels and evaluate the performance on various downstream tasks. Experiment results show that UniMoCo generalizes well for unsupervised, semi-supervised and supervised visual representation learning. The code is available: \url{https://github.com/dddzg/unimoco}.
\end{abstract}

\vspace{-4mm}
\section{Introduction}
 With well-designed pre-text tasks, unsupervised representation learning achieves great success in both natural language processing (NLP) and computer vision (CV). In NLP, the pre-text tasks are mainly designed to utilize the sequence relationship between discrete tokens (\eg GPT~\cite{radford2018improving,radford2019language} and BERT~\cite{devlin2018bert}). In CV, contrastive learning is proposed to learns visual representations from similar image pairs which are constructed by data augmentation (\eg MoCo~\cite{he2020momentum,chen2020improved} and SimCLR~\cite{chen2020simple}).
 
 \begin{figure}[t]
\centering
  \includegraphics[width=1.0\linewidth]{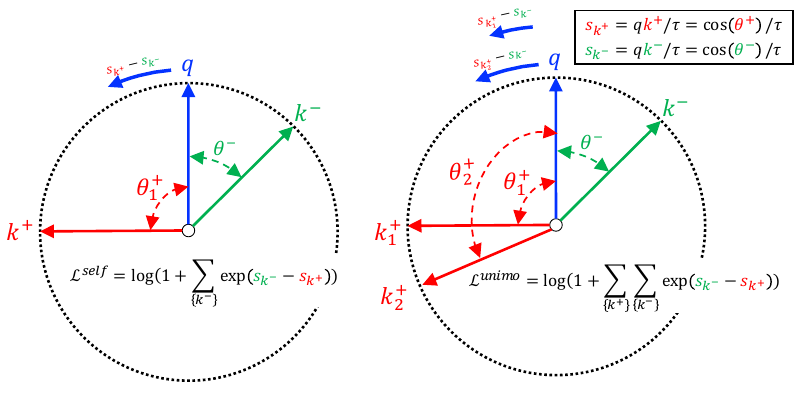}
  \small
  \begin{tabular}{cc}
  (a) MoCo\qquad&\qquad\qquad\qquad(b) UniMoCo
  \end{tabular}
  \caption{Illustrations of MoCo with $\mathcal{L}^{self}$ and UniMoCo with $\mathcal{L}^{unimo}$. In our perspective, these two losses both seek to reduce ($s_{k^-}-s_{k^+}$) among positives and negatives. For the query $q$, the loss pushes it close to positives $k^+$ and away from negatives $k^-$. (a) For MoCo, $\mathcal{L}^{self}$ only supports a single positive and multiple negatives. (b) For UniMoCo, $\mathcal{L}^{unimo}$ supports an arbitrary number of positives and negatives.}
  
  \label{intro}
  \vspace{-4mm} 
\end{figure}

In both NLP and CV, unsupervised representation learning transfers well to downstream tasks. However, there are significant differences in training datasets between NLP and CV on data labels and magnitude. For NLP, language models are pre-trained on a billion-word corpus, such as BooksCorpus~\cite{zhu2015aligning}, WebText~\cite{radford2019language} and English Wikipedia. The large-scale word corpus is scarcely possible to be labeled, so it is intuitive to develop completely unsupervised pre-training. For CV, most large-scale datasets, such as ImageNet~\cite{deng2009imagenet}, JFT~\cite{sun2017revisiting} and Instagram~\cite{mahajan2018exploring} are already labeled or partially labeled. Recent contrastive learning study~\cite{he2020momentum,chen2020improved,chen2020simple,chen2020exploring} simply discards all labels and outperforms supervised cross-entropy (CE) counterpart in downstream tasks. In \cite{khosla2020supervised}, Khosla \etal~ point out that supervised contrastive learning could also outperform the supervised cross-entropy counterpart. However, there is still lacking a quantitative evaluation about the label impact on contrastive learning in a unified framework.

 In this paper, we try to discuss this problem based on MoCo V2~\cite{chen2020improved}. However, MoCo only supports unsupervised learning with a single positive target by data augmentation. Therefore, we generalize MoCo from unsupervised learning into semi-supervised and full-supervised learning by two simple modifications:
 
\begin{enumerate}[leftmargin=1.5em]
\item[(1)] \textbf{Label queue.} To introduce label information, we maintain an extra label queue and update it with the feature queue together. Given the label of the query image, we can rapidly find out the same label samples in the queue and construct the multi-hot target on-the-fly.
\item[(2)] \textbf{Unified contrastive loss.} To support multiple positive  target, we generalize the original contrastive loss (InfoNCE)~\cite{oord2018representation} into a unified contrastive (UniCon) loss. In the further derivation~(\secref{compare_supcon}) and experiments~(\secref{abations}), the proposed UniCon loss is proved more reasonable and powerful than the SupCon~\cite{khosla2020supervised} losses. \figref{intro} shows the illustrations of MoCo with $\mathcal{L}^{self}$ and UniMoCo with $\mathcal{L}^{unimo}$.
\end{enumerate}
 
In our experiments, we train UniMoCo with different ratios of ImageNet labels from 0\% to 100\%, and evaluate the pre-training models on object detection, linear classification and network dissection~\cite{bau2017network}. UniMoCo is a powerful and universal framework to support unsupervised, semi-supervised and supervised visual representation learning. Experiment results show that: 
\begin{enumerate}[leftmargin=1.5em]
\item[(1)] \textbf{Contrastive loss is the kernel improvement for visual representation learning.} It performs better than CE counterpart in downstream tasks, regardless of unsupervised, semi-supervised or supervised learning.
\item[(2)] \textbf{More labels integrated into contrastive learning are beneficial for visual representation learning.} It not only sustains the linear classification accuracy as the CE counterpart, but also boosts the downstream task accuracy. 
\item[(3)] \textbf{The performance gain from full-supervision is limited.} Compared with semi-supervision of 60\% labels, full-supervision is only 0.2\% ImageNet linear classification top-1 accuracy improvement. For downstream detection tasks, full-supervision even performs a little worse than semi-supervision.

\end{enumerate}



\section{Related Work}
\subsection{Visual Representation Learning}
Before unsupervised pre-training, ImageNet supervised pre-training is dominant in computer vision, which is widely applied as the initialization for fine-tuning in downstream tasks~\cite{girshick2014rich,girshick2015fast,long2015fully,ren2015faster}. There are some consistent acknowledgments that ImageNet pre-training is a good initialization under controlled fine-tuning schedules~\cite{he2019rethinking} and better ImageNet models are correlated with better transfer performance~\cite{kornblith2019better}.

Recently, unsupervised visual representation learning~\cite{wu2018unsupervised,oord2018representation,hjelm2018learning,zhuang2019local,henaff2019data,tian2019contrastive,he2020momentum,chen2020improved,chen2020simple,grill2020bootstrap,chen2020exploring} with contrastive loss~\cite{hadsell2006dimensionality} achieves remarkable progress, which even outperforms supervised counterpart in lots of downstream tasks. Instance-based discrimination tasks~\cite{ye2019unsupervised,wu2018unsupervised} and clustering-based tasks~\cite{caron2018deep} are two typical pretext tasks for recent study.

Instance discrimination tasks are mainly various on maintaining the different sizes of negatives. For the original method~\cite{ye2019unsupervised,wu2018unsupervised}, there is a memory bank saving all image features from the dataset. The memory bank is large, however, features are inconsistent during the training. He \etal~\cite{he2020momentum} propose a momentum encoder with the queue to maintain large and consistent negative features. Chen \etal~\cite{chen2020simple} perform contrastive learning by straightforward large batch training end-to-end with the TPU support. Grill~\etal \cite{grill2020bootstrap} propose an approach with online and target networks, which even do not require any negatives. Chen~\etal~\cite{chen2020exploring} points out that a stop-gradient operation plays an essential role in preventing collapsing. 

Clustering-based tasks aim to learn visual representation and cluster together. DeepCluster~\cite{caron2018deep} is the typical method by iteratively updating model weights with clustering and label assignments. Asano~\etal~\cite{asano2019self} introduces the optimal transport problem with Sinkhorn-Knopp algorithm~\cite{cuturi2013sinkhorn} to balance the cluster and avoid degenerate solutions. Caron~\etal~\cite{caron2020unsupervised} propose an online algorithm, SwAV, which simultaneously clusters the data and performs contrastive learning with mini-batch training.
\begin{figure*}[ht]
\centering
\begin{subfigure}{0.45\linewidth}
\includegraphics[width=1.0\linewidth]{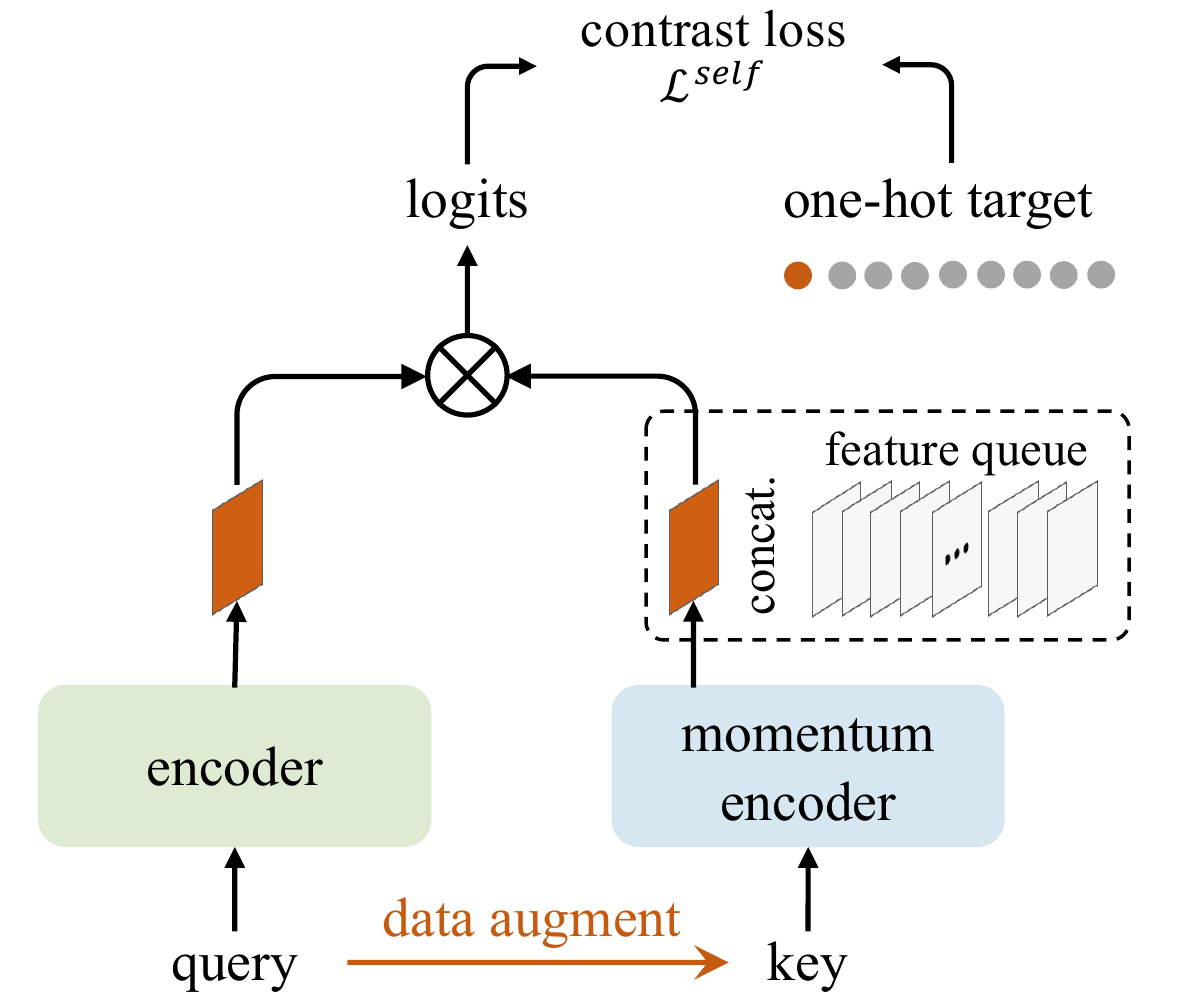} 
\caption{MoCo}
\label{fig:moco}
\end{subfigure}
\begin{subfigure}{0.45\linewidth}
\includegraphics[width=1\linewidth]{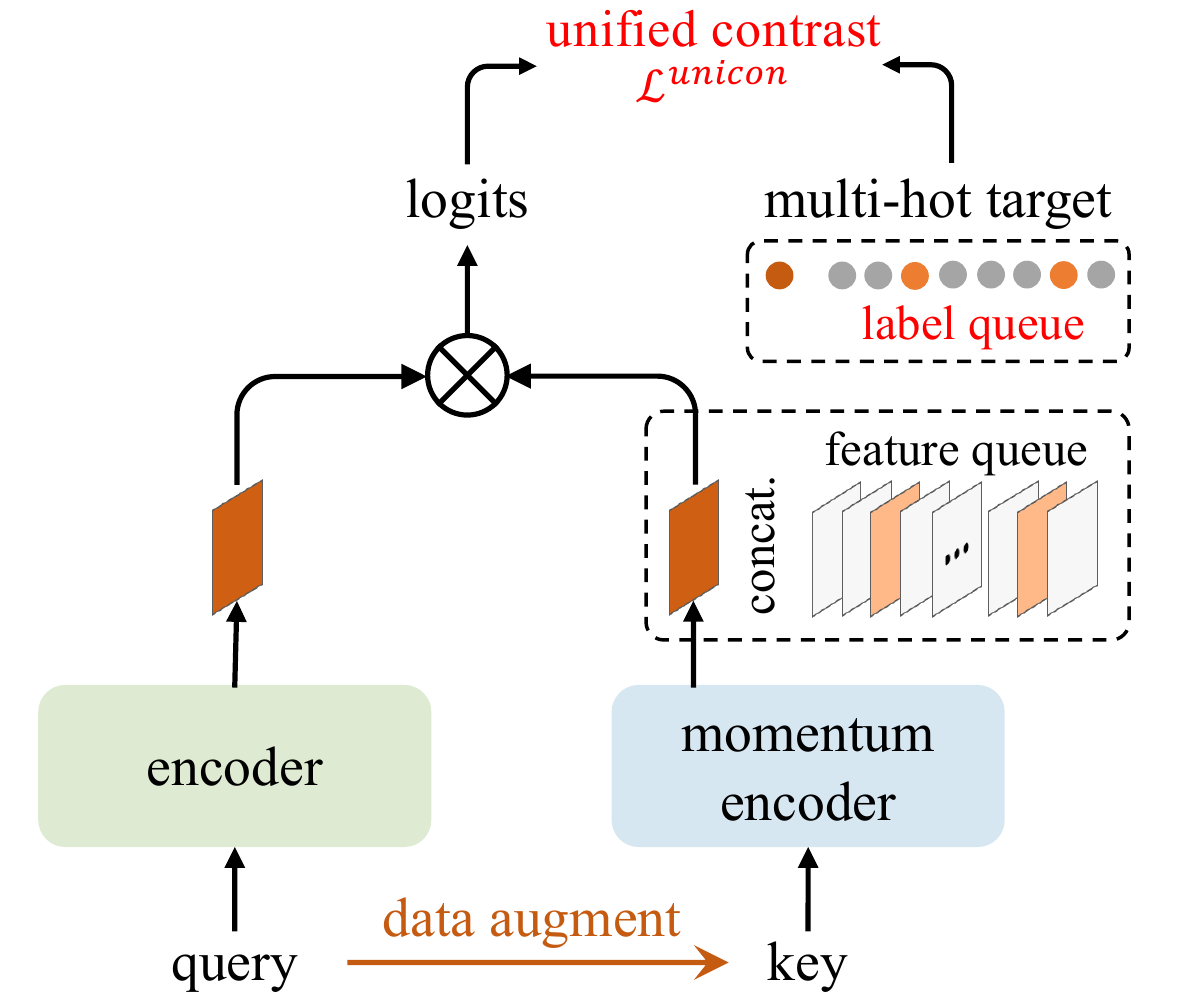}
\caption{UniMoCo}
\label{fig:unimoco}
\end{subfigure}
  \caption{Comparison of MoCo and UniMoCo. We annotate the main modifications with red and illustrate one pair of query and key for simplicity. (a): In MoCo, there is only a single positive key constructed from the data augmentation. So, the target is always a one-hot representation with $\mathcal{L}^{self}$. (b): In \textit{UniMoCo}, positives are consist of the data-augmented image and same-class images. The length of the label queue is the same as the feature queue. We compare the query label with the label queue to get the \textbf{multi-hot} target. $\mathcal{L}^{unico}$ is a unified loss that can deal with multiple positive pairs.}
  \label{fig:main}
  \vspace{-4mm}
\end{figure*}
\subsection{Loss Functions}
Generally, loss functions measure the distance between the model prediction and the fixed target. For example, the cross-entropy loss is a powerful and effective loss function to train deep networks~\cite{rumelhart1986learning,baum1987supervised} with the classification task. For cross-entropy, the target is organized as a one-hot representation with the class index, which is already pre-defined. However, cross-entropy also suffers from some drawbacks: sensitivity to noisy labels~\cite{zhang2018generalized,sukhbaatar2014training}, poor performance with adversarial examples~\cite{nar2019cross} and class-imbalance dataset~\cite{cao2019learning}. There are some popular solutions to replace the fixed target with non one-hot representation: \eg label smoothing to decrease the peakness with soft target~\cite{muller2019does,szegedy2016rethinking}, data augmentations with Mixup~\cite{zhang2017mixup}. Moreover, there are also some variants of cross-entropy in terms of the margin: \eg CosFace~\cite{wang2018cosface}, ArcFace~\cite{deng2019arcface} and Circle loss~\cite{sun2020circle}. They are widely used in the facial recognition system.

Different from cross-entropy, contrastive loss measures the similarities between different sample pairs with feature embedding interactions. With candidate sample pairs, the target of contrastive loss can vary during the training. For the contrastive loss, the logits are normalized by the estimation of sampled pairs, ~\eg noise-contrastive estimation (NCE)~\cite{gutmann2010noise}, instead of a summation of fixed classes in the cross-entropy.

Noting that Musgrave~\etal~\cite{musgrave2020metric} point out that the actual improvements of numerous metric learning methods are marginal at best under the fair comparisons. 

\vspace{-4mm}
\section{Method}

As shown in \figref{fig:main}, the main idea of UniMoCo is to generalize the MoCo framework from unsupervised representation learning to support semi-supervised and full-supervised learning. Therefore, We can further discuss the impact of label ratios on visual representation learning.


\subsection{MoCo}
For visual representation learning, MoCo~\cite{he2020momentum} learns the similar/dissimilar features from positive/negative pairs: the pair from the data augmentations of the same image construct to the positive pair and any two different images become the negative pair. To build a large and consistent dictionary, they maintain a feature queue, which is updated by a momentum-based moving average encoder. 

As shown in \figref{fig:moco}, the query image is encoded as $q$, and the corresponding positive and negative is encoded as $k^{+}$ and $k^{-}$, respectively. For simplicity, we notate the set of all samples in the momentum queue as $\{k\}={\{k^{+},k^{-}_{1}, k^{-}_{2}, ...\}}$. The self-supervised contrastive loss with the temperature $\tau$ is defined as:
\begin{align}
\mathcal{L}^{self} = -\log{\frac{\exp(q\small{\cdot} k^{+}/\tau)}{\sum\limits_{\{k\}}\exp(q\small{\cdot} k/\tau)}} = - \log{\frac{\exp(s_{k^+})}{\sum\limits_{\{k\}}\exp(s_{k})}},
\label{moco_loss}
\end{align}
where $s_k=q\small{\cdot}k/\tau$ is the logit for the given query $q$ and all the feature is normalized by the L2-norm.  Intuitively, \eqnref{moco_loss} can be treated as a log loss of single-label classification problem with $|\{k\}|$ classes, and the model is trained to find the unique positive pair $k^+$ from $\{k\}$.

\subsection{UniMoCo}
Supervised data can be treated as a set of human-defined positive pairs, and any two images with the same annotated label construct to the positive pair. Therefore, we can extend MoCo from the single positive pair to multiple positive pairs. However, MoCo does not maintain any label information and all the features in the queue are negatives. Moreover, the loss in \eqnref{moco_loss} can only support a single positive pair with multiple negative pairs. 
Generalizing from MoCo to UniMoCo, we propose two simple modifications including \textbf{label queue} and \textbf{unified 
contrastive loss} shown in \figref{fig:main}.

\subsubsection{Label Queue}
MoCo does not maintain any label information in the queue, where all samples are negatives and the only positive is transformed by data augmentation. For UniMoCo shown in \figref{fig:unimoco}, there are some potential positives with the annotated labels in the queue. To find the corresponding positives on-the-fly, we maintain a label queue and update it with feature queue together. Given a query $q$, we compare the query label with the label queue and get a multi-hot output as the target.

In particular, to be compatible with unlabeled data in unsupervised and semi-supervised situations, we label them as -1. During the training, we treat all samples with -1 label as negatives and preserving the augmented sample to positive. Therefore, UniMoCo is compatible with unsupervision, semi-supervision and full-supervision.


\vspace{-2mm}
\subsubsection{Unified Contrastive Loss}

$\mathcal{L}^{self}$ is designed for finding the single positive pair among all candidate pairs, which can be treated as a single-label classification problem. Multiple positive pairs can be treated as a multi-label classification problem. Khosla etal~\cite{khosla2020supervised} propose two versions of Supervised Contrastive (SupCon) loss to generalize \eqnref{moco_loss}. In this paper, we propose Unified Contrastive (UniCon) loss from a different perspective. Firstly, we rewrite the original contrastive loss $\mathcal{L}^{self}$ as follows:


\begin{align}
\mathcal{L}^{self} &= - \log{\frac{\exp(s_{k^+})}{\exp(s_{k^+})+\sum\limits_{\{k^-\}}\exp(s_{k^-})}} \\
&= \log(1+\sum\limits_{\{k^-\}}\exp(s_{k^-}-s_{k^+})),
\label{circle_loss_1}
\end{align}
where $\mathcal{L}^{self}$ seeks to reduce  $\exp(s_{k^-}-s_{k^+})$ between the single positive and all negatives. For extra positives, we can generalize \eqnref{circle_loss_1} by adding extra entries $\exp(s_{k^-}-s_{k^+})$ between positives and negatives. The UniCon loss can be defined as follows:
\begin{align}
\mathcal{L}^{unicon} &= \log(1+\sum\limits_{\{k^+\}}\sum\limits_{\{k^-\}}\exp(s_{k^-}-s_{k^+})) 
\label{circle_loss_4}
\end{align}



Here, we provide an understanding in terms of $\mathcal{L}^{unicon}$. As $\exp(\Delta)$ is a monotonically increasing function and $\exp(\Delta)$ is always positive, so we have the inequation as follows:
\begin{equation}
\begin{split}
\exp(\max(\{\Delta_i\}_{i=0}^n))\leq\sum_{i=0}^n\exp(\Delta_i)\leq\\(1+n)
\exp(\max(\{\Delta_i\}_{i=0}^n)),
\end{split}
\end{equation}
where $\Delta_0=0$ and $\Delta_i=s_{k^-}-s_{k^+}$. Then, we apply $\log()$ to the:
\begin{equation}
\begin{split}
\max(\{0,\Delta_1,\Delta_2,...,\Delta_n\})\leq\log(1+\sum_{i=1}^n\exp(\Delta_i))\leq \\
\max(\{0,\Delta_1,\Delta_2,...,\Delta_n\}) + \log(1+n).
\label{inequality_2}    
\end{split}
\end{equation}
According to \eqnref{inequality_2}, \eqnref{circle_loss_4} can be treated as a differentiable approximation to the maximum of $({s_{k^-}-s_{k^+}})$. During the training, the model seeks to minimise the maximum of $({s_{k^-}-s_{k^+}})$ among all the positives and negatives. The optimization goal is that all the entries ${s_{k^-}-s_{k^+}}\leq 0$, which is equivalent to  ${s_{k^-}}\leq s_{k^+}$. Therefore, $\mathcal{L}^{unicon}$ is a universal loss that pushes each positive logit $s_{k^+}$ larger than each negative logit $s_{k^-}$.

To simplify the code implementation, we rearrange the \eqnref{circle_loss_4} as follows:
\begin{align}
\mathcal{L}^{unicon} &= \log(1+\sum\limits_{\{k^-\}}\exp(s_{k^-})\sum\limits_{\{k^+\}}\exp(-s_{k^+})).
\label{circle_loss_final}
\end{align}
\algref{UniCon} shows the code of \eqnref{circle_loss_final} based on PyTorch~\cite{paszke2019pytorch}.


\begin{algorithm}[H]
\begin{lstlisting}[language=Python]
# logits and target with shape: Nx(1+K)
def UniCon(logits, target):
    sum_neg = ((1 - target) * torch.exp(logits)).sum(1)
    sum_pos = (target * torch.exp(-logits)).sum(1)
    loss = torch.log(1 + sum_neg * sum_pos)
    return torch.mean(loss)

\end{lstlisting}
\caption{Unicon Loss $\mathcal{L}^{unicon}$ based on PyTorch.}
\label{UniCon}
\end{algorithm}


\noindent\begin{minipage}{0.5\textwidth}
\renewcommand\footnoterule{}
\begin{algorithm}[H]
\begin{lstlisting}[language=Python]
# f_q, f_k: encoder networks for query and key
# f_que: feature queue of K keys (CxK)
# l_que: label queue of K keys (K)
# m: momentum, t: temperature
f_k.params = f_q.params # initialize
for x, <@\textcolor{red}{label}@> in loader:
    q = f_q.forward(aug(x)) # queries: NxC
    k = f_k.forward(aug(x)).detach() # keys: NxC
    logit_aug = bmm(q.view(N,1,C), k.view(N,C,1)) # Nx1
    logit_que = mm(q.view(N,C), f_que.view(C,K)) # NxK
    logits = cat([logit_aug, logit_que], dim=1) # Nx(1+K)
    
    # positive label for the augmented version
    <@\textcolor{red}{target\_aug = ones((N, 1))}@>
    # comparing the query label with l_que
    <@\textcolor{red}{target\_que = (label[:,None] == l\_queue[None,:])}@>
    # for the query labeled -1
    <@\textcolor{red}{target\_que \&= (label[:,None] != -1)}@>
    # labels: Nx(1+K)
    <@\textcolor{red}{target = cat([target\_aug, target\_que], dim=1)}@>
    # calculate the contrastive loss, Eqn.(7)
    <@\textcolor{red}{loss = UniCon(logits/t,target.float())}@>
    
    # update the query encoder and the key network
    loss.backward()
    update(f_q.params)
    f_k.params = m*f_k.params+(1-m)*f_q.params
    
    # update label queue and feature queue
    <@\textcolor{red}{enqueue\_and\_dequeue(l\_queue, label)}@>
    enqueue_and_dequeue(f_que, k)
\end{lstlisting}
\caption{Pseudocode of UniMoCo in PyTorch style.}
\label{uni_moco}
\end{algorithm}
\algcomment{The \textcolor[rgb]{1,0,0}{code with red} is the main differences to MoCo. \texttt{bmm}: batch matrix multiplication; \texttt{mm}: matrix multiplication; \texttt{cat}: concatenation.}
\end{minipage}

\subsubsection{Training}

During the training, we feed the query image $q$ with its label to the UniMoco framework.
Consistent with MoCo, we treat the augmented image as one of positive image by default for both unsupervised and supervised learning. Therefore, for the supervised data, positives come from two ways: (1) the augmented image from the same image and (2) images in the queue with the same label as the query. For the given query, we can get positives and negatives by comparing the $q$ label with the label queue. 


\algref{uni_moco} shows the pseudocode of UniMoCo in PyTorch~\cite{paszke2019pytorch} style. The \textcolor{red}{code with red} is the main differences to MoCo. Some variables are renamed due to the different meaning. 






\subsection{Comparison with SupCon Loss}
\label{compare_supcon}
Our method is functionally similar to supervised contrastive (SupCon)  loss~\cite{khosla2020supervised}. 
There are two formulas of SupCon loss defined as $\mathcal{L}^{sup}_{out}$ and $\mathcal{L}^{sup}_{in}$. These two SupCon losses are significantly different from the proposed UniCon loss. We discuss the differences between SupCon and UniCon as follows.

For the $\mathcal{L}^{sup}_{out}$, \eqnref{sup_out_1} calculates the InfoNCE loss of each positive pair, and then averaging them with the summation located outside the $\log$. In \eqnref{sup_out}, each positive $s_{k+}$ and each entries in $\{s_{k}\}$ contribute to $\exp(s_k - s_{k+})$ instead of $\exp(s_{k-} - s_{k+})$ in the UniCon loss. Therefore, $\mathcal{L}^{sup}_{out}$ takes same redundant entries $\exp(s_{k+}^i-s_{k+}^j)$ into account for multiple positives, and they are unnecessary for supervised contrastive learning. In additional, these redundant entries can not be canceled out due to the $\exp$ operator, such as $\exp(s_{k+}^1-s_{k+}^2)$ and $\exp(s_{k+}^2-s_{k+}^1)$. These redundant entries could be harmful to the model training. In \eqnref{circle_loss_4}, all entries of $\mathcal{L}^{unicon}$ are formed between positive pairs and negative pairs, which is more reasonable for contrastive learning~\footnote{Similar to $\mathcal{L}^{sup}_{out}$, we will discuss the other variant $\mathcal{L}^{unicon}_{out}$ in the appendix.}.
\begin{align}
\mathcal{L}^{sup}_{out} &= -\frac{1}{|\{k^+\}|}\sum\limits_{\{k^+\}}
\log(\frac{\exp(s_{k^+})}{\sum\limits_{\{k\}}\exp(s_{k})}) \label{sup_out_1}\\
&= \frac{1}{|\{k^+\}|}\sum\limits_{\{k^+\}}
\log(\sum\limits_{\{k\}}\exp(s_{k}-s_{k^+})),
\label{sup_out}
\end{align}

For the $\mathcal{L}^{sup}_{in}$, the summation in \eqnref{sup_in_1} is located inside the $\log$. The numerator is the summation of all positives, which can be treated that averaging the contribution of all positives into a positive $\frac{\sum_{\{k^+\}}\exp(s_{k^+})}{|\{k^+\}|}$ and then calculating the InfoNCE loss. A potential problem of $\mathcal{L}^{sup}_{in}$ is that, the denominator is still $\sum_{\{k\}}\exp(s_k)$, which count all candidate samples. Logically, it should be $\sum_{\{k^-\}}\exp(s_{k^-})+\frac{\sum_{\{k^+\}}\exp(s_{k^+})}{|\{k^+\}|}$ (summation with negatives and the averaged positive) following the InfoNCE loss. In other words, it only average positives in the numerator, but not averaging them in the denominator. Therefore, it may lead to a biased statistic.
\begin{align}
\mathcal{L}^{sup}_{in} &= -\log(\frac{1}{|\{k^+\}|}\sum\limits_{\{k^+\}} \frac{\exp(s_{k^+})}{\sum\limits_{\{k\}}\exp(s_{k})}) \label{sup_in_1}\\
&= -\log(\frac{\sum_{\{k^+\}}\exp(s_{k^+})}{|\{k^+\}|\sum_{\{k\}}\exp(s_{k})}).
\label{sup_in}
\end{align}

In summary, when there is only a single positive pair $s_{k+}$, $\mathcal{L}^{sup}_{out} = \mathcal{L}^{sup}_{out} = \mathcal{L}^{unicon}$. They are all straightforward extensions to generalize $\mathcal{L}^{self}$. However, these two versions of SupCon losses still follow the softmax form, averaging all entries into the denominator, which leads to a biased statistic of the denominator. However, $\mathcal{L}^{unicon}$ is developed in the perspective of pair-wise optimization with the entry ($s_{k^-}-s_{k^+}$). Therefore, our $\mathcal{L}^{unicon}$ is more reasonable to generalize $\mathcal{L}^{self}$ with a symmetry form.

\begin{table*}[t]
\begin{tabular}{ll|lll|lll}
Pre-train& ratio & $AP^{bb}$   & $\textrm{AP}^{bb}_{50}$ & $\textrm{AP}^{bb}_{75}$ & $AP^{mk}$ & $\textrm{AP}^{mk}_{50}$ & $\textrm{AP}^{mk}_{75}$ \\\thickhline
\gray{Rand. init.}& 0\% &  \gray{26.4} & \gray{44.0} & \gray{27.8} & \gray{29.3} & \gray{46.9} & \gray{30.8} \\\hline
MoCo v1 & 0\% & 38.5 & 58.3 & 41.6 & 33.6 & 54.8 & 35.6 \\
SimCLR & 0\% & 37.9 & 57.7 & 40.9 & 33.3 & 54.6 & 35.3 \\
MoCo v2 & 0\% &  39.0 & 58.6 & 42.0 & 34.1 & 55.3 & 36.2 \\
BYOL & 0\% & 37.9 & 57.8 & 40.9 & 33.2 & 54.3 & 35.0 \\
SwAV & 0\% & 37.6 & 57.6 & 40.3 & 33.1 & 54.2 & 35.1 \\
SimSiam & 0\% & 39.2 & 59.3 & 42.1 & 34.4 & 56.0 & 36.7 \\
CE Super.& 100\%  &  38.2 & 58.2 & 41.2 & 33.3 &54.7 & 35.2 \\\hline
UniMoCo& 0\% &  39.0 & 58.6 & 42.0 & 34.1 & 55.3 & 36.2 \\
UniMoCo& 10\% &  39.3 & 59.0 &	42.5 &	34.5 &	56.0 &	36.8 \\
UniMoCo& 30\% &  39.6 & 59.4 &	42.6 & \textbf{34.6} & \textbf{56.1} & 36.8 \\
UniMoCo& 60\% &  \textbf{39.8} & \textbf{59.5} &	\textbf{43.2} &	\textbf{34.6} &	\textbf{56.1}  &	\textbf{37.0}  \\
UniMoCo& 100\% & 39.6 & 59.3 &	42.5 &	\textbf{34.6} &	\textbf{56.1}  &	36.7 
 \\
 \multicolumn{7}{c}{(a) Mask R-CNN, R50-\textbf{C4},\textbf{1}$\times$ schedule}
\end{tabular}~
\begin{tabular}{lll|lll}
 $AP^{bb}$   & $\textrm{AP}^{bb}_{50}$ & $\textrm{AP}^{bb}_{75}$ & $AP^{mk}$ & $\textrm{AP}^{mk}_{50}$ & $\textrm{AP}^{mk}_{75}$ \\\thickhline
 \gray{35.6} & \gray{54.6} & \gray{38.2} & \gray{31.4} & \gray{51.5} & \gray{33.5} \\\hline
 40.7 & 60.5 & 44.1 & 35.4 & 57.3 & 37.6 \\ 
 - & - & - & - & - & - \\
 40.8 & 60.5 & 44.5 & 35.5 & 57.5 & 38.0 \\
 - & - & - & - & - & - \\
 - & - & - & - & - & - \\
 - & - & - & - & - & - \\
 40.0 & 59.9 & 43.1 & 34.7 & 56.5 & 36.9 \\\hline
  40.8 & 60.5 & 44.5 & 35.5 & 57.5 & 38.0 \\
 40.9 &	60.7 & 44.5 & 35.6 & 57.4 &	38.0 \\
 40.9 &	60.8 &	44.4 &	35.6 &	57.4 &	37.9  \\
 \textbf{41.3} &	\textbf{61.1} &	44.7 &	\textbf{35.9} &	\textbf{57.8} & \textbf{38.3}
 \\
 \textbf{41.3} &	\textbf{61.1} & \textbf{44.9} & 35.7 & 57.6 &	38.1 \\
\multicolumn{6}{c}{(b) Mask R-CNN, R50-\textbf{C4},\textbf{2}$\times$ schedule}
\end{tabular}
\caption{\textbf{Object detection and instance segmentation fine-tuned on COCO}~\texttt{train2017} \textbf{and evaluated on}~\texttt{val2017}. All pre-training models are pre-trained on ImageNet with the R50-C4 backbone. $\textrm{AP}^{bb}$ and $\textrm{AP}^{mk}$ refer to bounding box AP and mask AP of Mask R-CNN with 1$\times$ and 2$\times$ schedule. For the supervised model trained with cross-entropy loss, we name it CE Super. (100\%).}
\label{coco_exp}
\end{table*}

\section{Experiments}
We pre-train standard ResNet-50~\cite{he2016deep} models on 1.28M training data of ImageNet~\cite{deng2009imagenet} with different ratios of labels from unsupervised, semi-supervised to supervised: 0\%, 10\%, 30\%, 60\%, 100\%. We name our models as UniMoCo (label ratio) in our experiments. The supervised model trained by cross-entropy loss is named CE Supervised (100\%). It is worth noting that we develop our method based on MoCo v2 open-source codebase~\footnote{https://github.com/facebookresearch/moco} with the same setting, which leads to the fact that UniMoCo(0\%) is \textbf{exactly same as} MoCo v2 in theory. Therefore, we report the same MoCo v2 precision for UniMoCo (0\%) in the following experiments without re-run.
All UniMoCo models are pre-trained for 800 epochs with 256 batch size in 8 GPUs by default in main experiments. For the label ratio $\alpha$, the size of the queue $K$ and the number of classes is $C$ (1000 for ImageNet), there is about $\alpha*K/C$ positives in the queue. We adopt $K=65536$ and $\alpha=0, 0.1, 0.3, 0.6, 1$ to represent different magnitude supervised information. To discuss the label impact to the representation learning, we conduct three evaluations: fine-tuning pre-training weights on COCO~\cite{lin2014microsoft} and PASCAL VOC~\cite{everingham2010pascal}, training a linear classification on frozen features and performing network dissection~\cite{bau2017network}. It is worth noting that, we use 2 layers in the projection MLP with the asymmetric loss for all pre-trained models. All the optimizers are SGD with 0.9 momentum. So, the baseline model UniMoCo (0\%) is a little lower than the MoCo v2 implemented in SimSiam ~\cite{chen2020exploring}.

\subsection{COCO Object Detection}

\tabref{coco_exp} shows the result fine-tuned on COCO \texttt{train2017} under the same setup as MoCo open-source codebase. With 1$\times$ training schedule, UniMoCo (60\%) performs best, which outperforms CE Supervised (100\%) and UniMoCo (0\%) by up to 1.6 and 0.8 points AP$^{bb}$, respectively. With 2$\times$ training schedule, UniMoCo (60\%) still performs best, and it outperforms CE Supervised (100\%) and UniMoCo (0\%) by up to 1.3 and 1.1 points AP$^{bb}$, respectively. The trend of AP$^{mk}$ is consistent with AP$^{bb}$. Overall, when UniMoCo pre-trained with different ImageNet label ratios, AP$^{bb}$ and AP$^{mk}$ increase as the labels ratio increase. It also suggests that contrastive learning can still benefit from annotated labels pre-training. Moreover, UniMoCo (100\%) performs a little worse than UniMoCo (60\%) about 0.2\% AP$^{bb}$ and AP$^{mk}$, which suggests that partial labels on ImageNet may not beneficial for transfer learning. For 100\% labels, UniMoCo (100\%) performs much better than CE Super. (100\%). Overall, models trained with contrastive loss, regardless of unsupervised, semi-supervised or supervised learning, is better than the CE supervised counterpart.



\begin{table}[t]
\centering
\begin{tabular}{ll|l|ll}
Pre-train & ratio & $\textrm{AP}_{50}$ & AP & $\textrm{AP}_{75}$ \\\thickhline
\gray{Random init.} & \gray{0\%} & \gray{64.4} & \gray{37.9} & \gray{38.6} \\ \hline
MoCo v1~\cite{ye2019unsupervised} & 0\% & 81.5 & 55.9 & 62.6 \\
SimCLR~\cite{chen2020simple} & 0\% & 81.8 & 55.5 & 61.4 \\
MoCo v2~\cite{chen2020improved}& 0\% & 82.5 & 57.4 & 64.0  \\
BYOL~\cite{grill2020bootstrap} & 0\% & 81.4 & 55.3 & 61.1 \\
SwAV~\cite{caron2020unsupervised}& 0\% & 81.5 & 55.4 & 61.4 \\
SimSiam~\cite{chen2020exploring}& 0\% & 82.4 & 57.0 & 63.7 \\
CE Super. & 100\% & 81.3 & 53.5 & 58.8 \\ \hline
UniMoCo & 0\%  & 82.5 & 57.4 & 64.0  \\
UniMoCo & 10\%  &  82.6 & 57.4 & 64.1 \\
UniMoCo & 30\%  &  82.8 & 57.7 & 64.3 \\
UniMoCo & 60\%  &  \textbf{82.9} & \textbf{57.8} & \textbf{64.7} \\
UniMoCo & 100\% &  82.8 & \textbf{57.8} & 64.6 \\
 
\end{tabular}
\caption{\textbf{Object detection fine-tuned on PASCAL VOC} \texttt{trainval07+12} \textbf{and evaluated on} \texttt{test2007}. All pre-training models are pre-trained on ImageNet with the R50-C4 backbone. All metrics: $\textrm{AP}_{50}$(default VOC metric), AP, $\textrm{AP}_{75}$ are averaged over 5 trials. For the supervised model trained with cross-entropy loss, we name it CE Super. (100\%).}
\label{voc_exp}
\end{table}

\subsection{PASCAL VOC Object Detection}
\label{voc}

\tabref{voc_exp} shows the result fine-tuned on PASCAL VOC \texttt{trainval07+12} under the same MoCo v2 setup. The overall trend of AP on VOC is similar to COCO. For UniMoCo models, AP increases as the label ratios increase. It achieves best pre-trained with 60\% labels. Although the numerical gains are limited (about 0.4 AP$_{50}$ and 0.4 AP), the improvement trend is obvious. Compared with unsupervised contrastive learning, adding supervised labels into pre-training UniMoCo improves the fine-tuning performance on VOC. Compared with supervised learning with 100\% labels, UniMoCo (100\%) surpasses it in all metrics with large improvements (1.6 AP$_{50}$ and 4.3 AP). It further verifies that the cross-entropy is the limitation of supervised visual representation learning.

\begin{table}[t]
\centering
\begin{tabular}{ll|l|r}
Method& ratio & Top-1 acc & Top-5 acc \\\thickhline

MoCo v1~\cite{ye2019unsupervised} & 0\% & 60.6 & - \\
SimCLR~\cite{chen2020simple} & 0\% & 69.3 &  89.0 \\
MoCo v2~\cite{chen2020improved} & 0\% & 71.1 & 90.1  \\
BYOL~\cite{grill2020bootstrap} & 0\% & 74.3 & 91.6 \\
SwAV~\cite{caron2020unsupervised} & 0\% & 75.3 & - \\
SimSiam~\cite{chen2020exploring} & 0\% & 71.3 & - \\ 
CE Super. & 100\% & \textbf{76.5} & \textbf{93.1} \\ \hline
UniMoCo & 0\% & 71.1 & 90.1  \\
UniMoCo & 10\% & 72.0 & 90.3 \\
UniMoCo & 30\% & 75.1 & 92.4 \\
UniMoCo & 60\% & 76.2 & 93.0 \\
UniMoCo & 100\% & \textbf{76.4} & \textbf{93.1} 
 
\end{tabular}
\caption{\textbf{Comparison under the linear classification protocol with ResNet-50 on ImageNet.} The upper part of the table is the models pre-trained with unsupervised contrastive learning pre-training and CE Super.(100\%). }
\label{linear_cls}
\end{table}

\begin{figure*}[]
\centering
\begin{subfigure}{0.48\linewidth}
\includegraphics[width=0.96\linewidth]{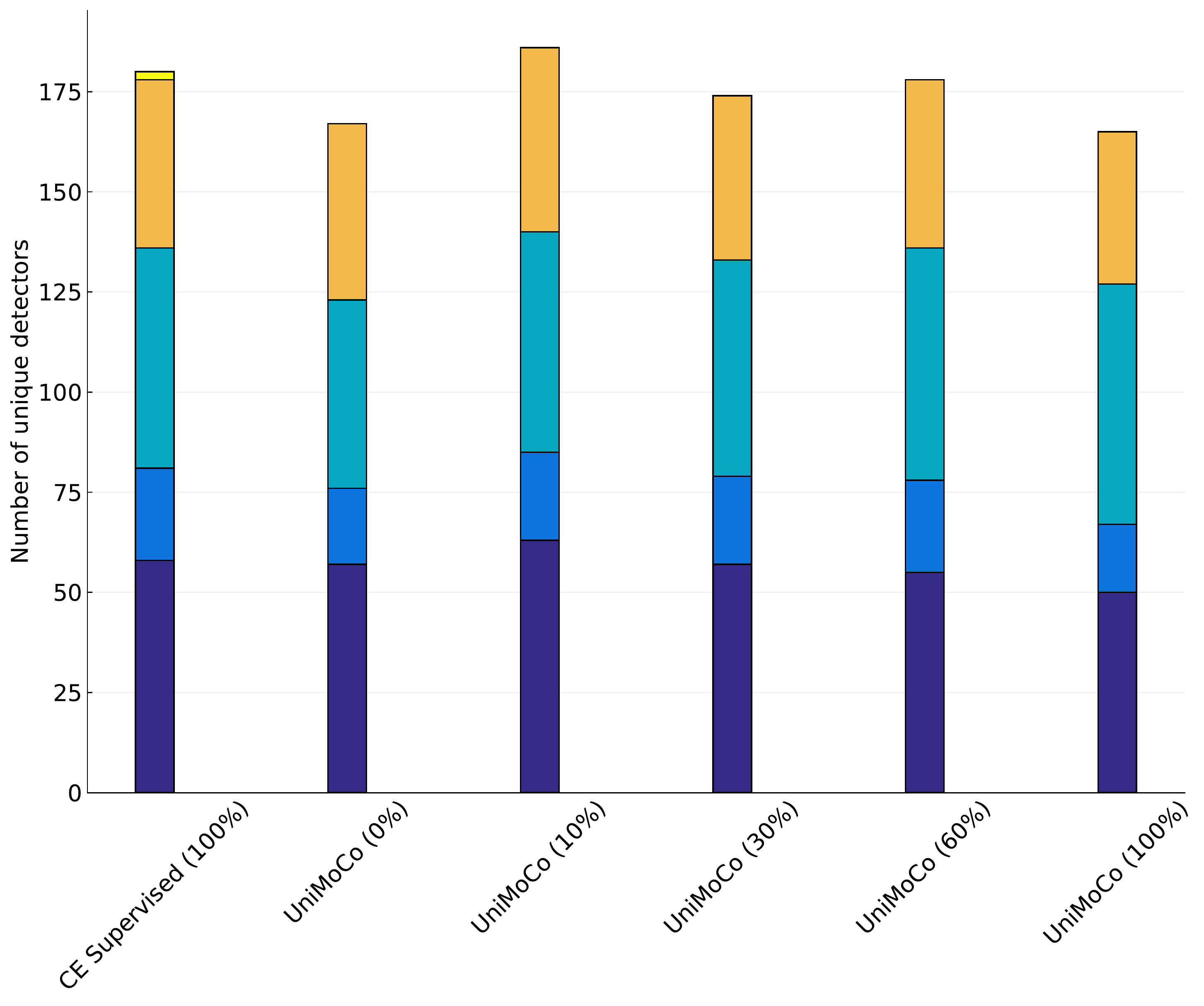} 
\caption{Number of unique detectors}
\label{fig:uni_det}
\end{subfigure}
\begin{subfigure}{0.48\linewidth}
\includegraphics[width=0.96\linewidth]{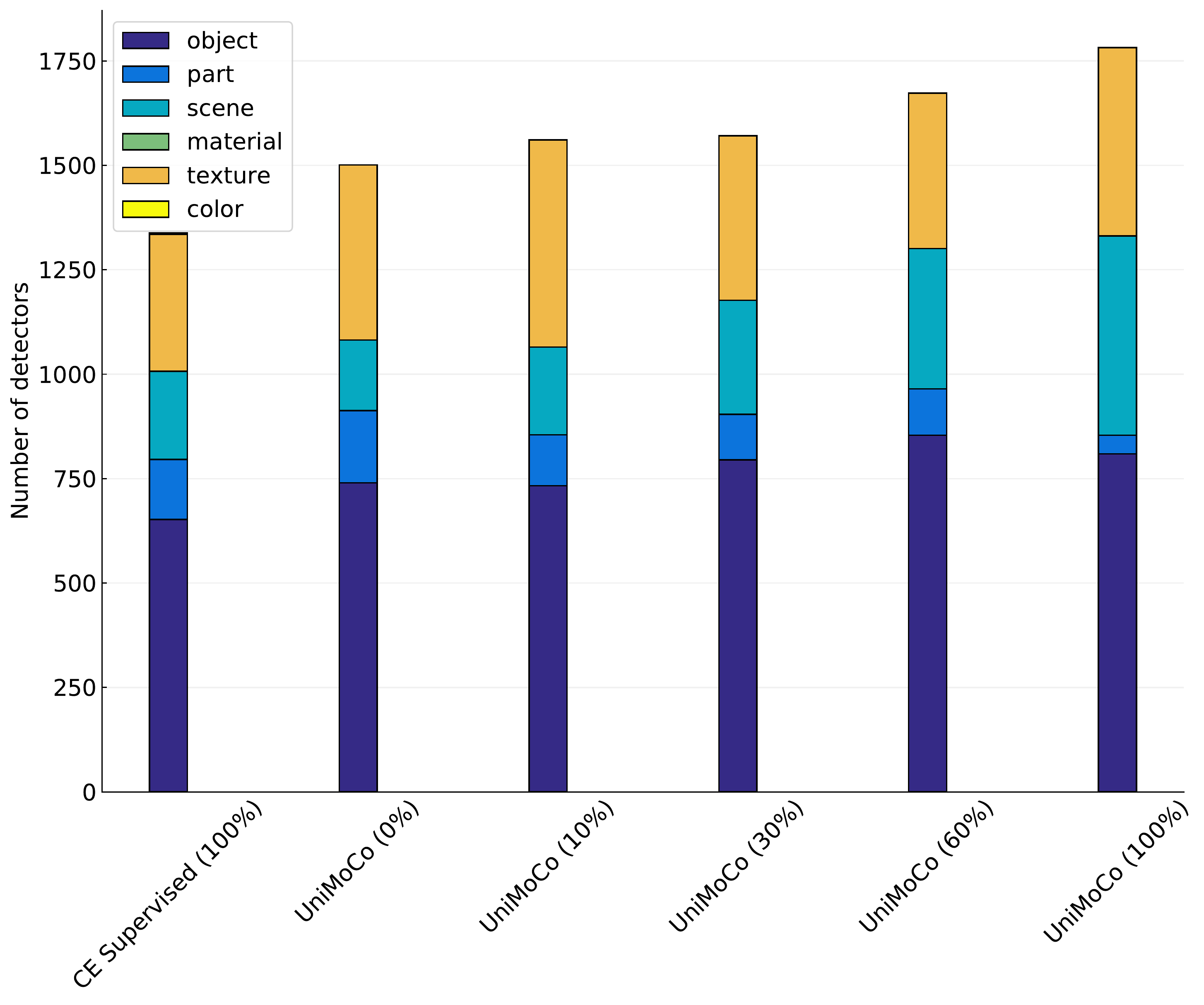}
\caption{Number of detectors}
\label{fig:det}
\end{subfigure}

\caption{Network dissection results of different pre-training ResNet-50 models at \texttt{conv5}. We follow the same six categories adopted in ~\cite{bau2017network,zhou2018interpreting}: object, scene, part, material, texture and color. }
\label{fig:net_dis}
\end{figure*}

\subsection{Linear Classification Protocol}
\noindent\textbf{Setup.} To verify pre-trained models are learned from supervised labels, we perform the linear classification on frozen features with ResNet-50, following a common protocol. For UniMoCo pre-trained with different labels, we train an extra linear classifier with 100 epochs on the top of the frozen features after the global average pooling~\cite{lin2013network}. Then, we evaluate the top-1 and top-5 classification accuracy on the ImageNet validation set with a single center crop (224 $\times$ 224). We train the linear layer with SGD optimizer following ~\cite{he2020momentum} with base $lr$ = 30.0, weight decay = 0, momentum = 0.9, and batch size=256.


\noindent\textbf{Result.} \tabref{linear_cls} shows the top-1 accuracy of linear classification protocol with different pre-training models. We observe that the top-1 accuracy of UniMoCo increases from 71.1 to 76.4 as the label ratio increases. It suggests that UniCon loss is an effective loss to boost the feature discrimination for unsupervised, semi-supervised and supervised learning. In particular, UniMoCo (100\%) obtains \textbf{76.4} ImageNet top-1 accuracy, which is comparable to CE Supervised (100\%) with \textbf{76.5}. they get the same top-5 accuracy. It further demonstrates that UniCon loss not only improves the performance in transfer learning, but also sustains the feature discrimination as CE loss. With enough training schedule, the feature discrimination of UniMoCo (100\%) is almost the same as the CE supervised counterpart. In other words, supervised contrastive learning needs more training iterations than CE supervised counterpart.


It is worth noting that Khosla \etal~\cite{khosla2020supervised} find that their supervised contrastive loss could surpass supervised cross-entropy loss with carefully chosen hyper-parameters and optimizer. For example, their best result is using AutoAugment~\cite{cubuk2018autoaugment} with LARC optimizer~\cite{you2017large}. For fair comparisons, we discuss the comparison with supervised contrastive loss in ~\secref{abations}.

\subsection{Network Dissection}
\noindent\textbf{Setup.}  Besides transferring features to downstream object detection task and linear classification protocol, we also perform the experiments with Network Dissection~\cite{bau2017network,zhou2018interpreting} to interpret visual representations without fine-tuning and re-train the extra classification head. Fine-tuning on downstream tasks and perform linear classification will require a lot of resources to verify the generalization of pre-training models. Instead, network dissection is a more effective and efficient method to evaluate the generalization and interpretability of pre-training models. For ResNet-50, the evaluation can be done within 1.5 hours on 1 GPU. In the following experiments, we report the number of unique detectors and the number of detectors at \texttt{conv5} for a thorough evaluation.


\noindent\textbf{Result.} \figref{fig:uni_det} shows the number of unique detectors (the default metric in network dissection) and number of detectors. As we can see, unique semantic detectors emerge most in UniMoCo (10\%) and it indicates that there is not a clear trend between the number of unique detectors and label ratios. But, at least, we can find that supervision is not a determining factor in the number of unique detectors. A reasonable explanation we notice is that one unit might detect multiple concepts at \texttt{conv5} of ResNet-50. Some concepts are easily detected with large IoU. However, in network dissection, they only count each unit once with the largest IoU ranked concept, which will ignore some concept detectors. It suggests that there may be a better evaluation to interpret the network interpretability, which needs to study in the future.


We notice that the result of number of detectors, shown in \figref{fig:det}, is more consistent with the result of downstream transfer performance. The number of detectors increases consistently as the ratio of labels increases. And, the number of detectors of UniMoCo(0\%) is more than CE supervised(100\%). The result suggests that the number of detectors may be a good metric to evaluate the model generalization. It is reasonable that more semantic detectors could provide more possibilities during transfer learning.

At least, we can conclude that supervision labels are not critical to network interpretability. There are more detectors emerging during the training of UniMoCo (100\%) than CE Supervised (100\%) and UniMoCo (0\%). Generally, the number of detectors increases as the ratio of labels increases for UniMoCo.


\begin{table}[t]
\centering
\begin{tabular}{l|l|lll}

 Loss & Top-1 acc. & Top-5 acc. \\\thickhline
 SupCon (inside) & 72.4 & 91.1 \\
 SupCon (outside) & 73.4 & 91.7 \\
 \textbf{UniCon (ours)} & \textbf{74.6} &  \textbf{92.2}
 
\end{tabular}
\caption{Comparisons of linear classification protocol for different UniMoCo with 100\% labels. All UniMoCo are pre-trained for 200 epochs. }
\label{tb_ablations}
\end{table}

\subsection{Comparison with Supervised Contrastive Loss}
\label{abations}
To eliminate the misjudgment of different hyper-parameters, we implement two versions of SupCon losses in our UniMoCo (100\%) framework with the same epochs and batch size training. We use the same data augmentation following \cite{chen2020improved}. UniMoCo (100\%) models with different loss functions are pre-trained for 200 epochs. In \tabref{tb_ablations}, we report ImageNet linear classification with Top-1 and Top-5 accuracy on the frozen feature for these three models. Our UniCon loss achieves 74.6\% ImageNet top-1 accuracy, which is 1.2\% and 2.2\% higher than SupCon (outside) and SupCon (inside), respectively. It verifies that our UniCon loss does perform better than SupCon in practice.

For the two versions of SupCon loss, outside loss performs better than inside loss. This conclusion is consistent with Khosla ~\etal~\cite{khosla2020supervised}. However, the gap between them in our UniMoCo framework is small (about 1\%), which is very large (about 11.3\%) in their original paper report. It suggests that the large queue (typically for 65536 samples) is better than 6144 batch samples training, which mitigates the statistics bias of the denominator in \eqnref{sup_in}.



\label{Analysis}

\section{Discussions}


We present UniMoCo with the UniCon loss for unsupervised, semi-supervised and full-supervised presentation learning. With UniMoCo, we discuss the label impact on contrastive learning. Results show that, for current contrastive learning, human annotations still help representation learning. Although unsupervised visual representation learning attracts a lot of research interests, which even surpasses the CE supervised counterpart. Actually, the improvement of visual representation learning comes from contrastive loss instead of unsupervised learning itself. Moreover, supervised learning with contrastive loss boosts the performance of visual representation in transfer learning and linear classification protocol. In other words, ImageNet labels are still beneficial for contrastive learning. It also means that unsupervised learning does not surpass supervised learning with the same contrastive loss. As far as we observe, Compared with semi-supervision, the performance gain from full-supervision is limited. It suggests that partial ImageNet labels may not beneficial for visual representation learning.

As more and more powerful representation learning algorithms are proposed, we hope our study could attract further study to rethink the importance of labels. Actually, there is potential competition between human annotations and algorithms. We are looking forward to a method and theory to evaluate the limitations of algorithms and human annotations objectively.


{\small
\bibliographystyle{ieee_fullname}
\bibliography{egbib}
}

\newpage
\section{Appendix}
\subsection{Comparison with Triplet Loss}
In our perspective, triplet loss ~\cite{weinberger2009distance} can be treated as a special case when there is only a positive and negative with margin = 0. for the L2-normalized feature $q,k^+,k^-$, triplet loss is defined as:
\begin{align}
\mathcal{L}^{triplet} &= \max(0, \left\|p-k^-\right\|^2-\left\|p-k^+\right\|^2) \\
&= 2 \max(0, p\small{\cdot} k^{-}-p\small{\cdot} k^{+}) \\
\intertext{As $s_k=q\small{\cdot}k/\tau$:}
&= 2\tau \max(0, s_{k^{-}}-s_{k^{+}}).
\end{align}
For our $\mathcal{L}^{unico}$, it is a differentiable approximation to $\max(0, s_{k^{-}}-s_{k^{+}})$ among all negatives and positives. So, $\mathcal{L}^{triplet} \approx 2\tau\mathcal{L}^{unico}$ when there is a single positive and negative.

\subsection{The other extension of UniCon Loss}
For $\mathcal{L}^{unico}$, the summation is located inside the $\log$. Similar to $\mathcal{L}^{sup}_{out}$, we can also extend $\mathcal{L}^{unico}$ to $\mathcal{L}^{unico}_{out}$ as follows:
\begin{equation}
\begin{split}
\mathcal{L}^{unico}_{out} = \frac{1}{|\{k^+\}|}\sum\limits_{\{k^+\}}
\log(1 + \sum\limits_{\{k^-\}}\exp(s_{k^-}-s_{k^+})). 
\end{split}
\end{equation}
As the differentiable approximation of $\max$ , $\mathcal{L}^{unico}_{out}$ can be treated as $\frac{1}{|\{k^+\}|}\sum\limits_{\{k^+\}}\max(0,s_{k^-}-s_{k^+})$. It means that for the given $s_{k^+}$, it compute the maximum of $({s_{k^-}-s_{k^+}})$ of each positive among all negatives, then averaging them.

However, compared with $\mathcal{L}^{unico}_{out}$, $\mathcal{L}^{unico}$ has two merits:

\begin{enumerate}[leftmargin=1.5em]
\item[(1)] \textbf{$\mathcal{L}^{unico}$ is more symmetrical.} For $\mathcal{L}^{unico}$, the summation of positives is located inside the $\log$ with negatives together. It is more unified and straightforward.
\item[(2)] \textbf{$\mathcal{L}^{unico}$ is beneficial for hard sample mining.} Hypothesise that there are a lot of easy positives and negatives. For the hard sample, easy samples will weaken the loss due to the average in the $\mathcal{L}^{unico}_{out}$.
\end{enumerate}

\end{document}